\pdfoutput=1

\documentclass[11pt]{article}
\usepackage{graphicx}
\usepackage{subfig}
\usepackage{appendix}
\usepackage[final]{acl}

\usepackage{times}
\usepackage{latexsym}
\usepackage{amsmath}
\usepackage{float}
\usepackage[T1]{fontenc}

\usepackage[utf8]{inputenc}

\usepackage{microtype}

\usepackage{inconsolata}

\usepackage{graphicx}

\title{Evaluating Automatic Metrics \\ with Incremental Machine Translation Systems}

\author{
 \textbf{Guojun Wu}\textsuperscript{\textnormal{1}} \quad
 \textbf{Shay B. Cohen}\textsuperscript{\textnormal{2}} \quad
 \textbf{Rico Sennrich}\textsuperscript{\textnormal{1,2}} \\
 \textsuperscript{1}Department of Computational Linguistics, University of Zurich \\
 \textsuperscript{2}School of Informatics, University of Edinburgh \\
 \small{\texttt{guojun.wu@uzh.ch,} \texttt{scohen@inf.ed.ac.uk,} \texttt{sennrich@cl.uzh.ch}}
}

\begin{document}
\maketitle
\begin{abstract}
We introduce a dataset comprising commercial machine translations, gathered weekly over six years across 12 translation directions. Since human A/B testing is commonly used, we assume commercial systems improve over time, which enables us to evaluate machine translation (MT) metrics based on their preference for more recent translations. Our study not only confirms several prior findings, such as the advantage of neural metrics over non-neural ones, but also explores the debated issue of how MT quality affects metric reliability—an investigation that smaller datasets in previous research could not sufficiently explore. Overall, our research demonstrates the dataset's value as a testbed for metric evaluation. We release our code.\footnote{\url{https://github.com/gjwubyron/Evo}}
\end{abstract}

\section{Introduction}
\label{sec:intro}

Automatic metrics for machine translation (MT) are typically assessed by measuring their correlation with or accuracy with respect to human judgments \cite{machacek-bojar-2013-results,mathur-etal-2020-results,kocmi-etal-2021-ship}. However, human evaluation is resource-intensive and time-consuming, and the number of translation systems included in a meta-evaluation tends to be relatively small. In this study, we explore the use of commercial machine translations, collected weekly over a period of 6 years for 12 translation directions, for the evaluation of MT metrics. Given the common use of human A/B testing \cite{abtesting,caswell2020google}, our base assumption is that commercial systems show real improvements over time and that we can assess metrics as to whether they prefer more recent MT outputs. Using our dataset, we revisit a number of recent findings in MT metrics research, and find that our dataset supports these. 

Trained metrics, developed to directly learn human judgments \cite{rei-etal-2020-comet, sellam-etal-2020-bleurt}, showed notable advancements in correlating with human judgments compared to non-neural metrics like BLEU \cite{papineni-etal-2002-bleu,freitag-etal-2021-results}. Recent studies \cite{freitag-etal-2022-results,freitag-etal-2023-results} also revealed that neural metrics achieved significantly better correlation with human judgments than non-neural ones and were able to generalize to new domains and challenging test sets. In our experiments, we analyze metric scores over time and evaluate metrics' ability to accurately rank MT systems. Our findings demonstrate that neural metrics show a more consistent upward trend, and achieve higher accuracy than non-neural metrics.

\citet{ma-etal-2019-results} argued that the correlation between metrics and human judgments significantly decreased when considering only the top-performing systems. To be specific, they assessed the stability of metrics across top-N MT systems, and noticed that the correlation between metric and human scores diminished as N decreased. \citet{mathur-etal-2020-tangled} highlighted the instability of correlations with small N, and instead employed a rolling window of N systems, moving from the worst to the best systems. Due to the limited number of MT systems (typically 10-15 systems) in the datasets, they were unable to confirm that the correlation declines as system quality improves with their approach. Due to the much larger size of our dataset (see Section~\ref{sec:data}), we can achieve more stable results when using the rolling window approach from \citet{mathur-etal-2020-tangled}. Our findings reveal that a downward trend is the most common, supporting the results of \citet{ma-etal-2019-results}, although upward or relatively flat trends are also seen in some language pairs.

In WMT23 Metrics shared task \cite{freitag-etal-2023-results}, human translations received unexpectedly low ratings, which prompted an investigation into using synthetic references as a potential alternative. They found that high-quality synthetic references could produce a stronger correlation between human evaluations and metrics compared to human references. In our study, we reexamine the usefulness of synthetic references with three language pairs and find that synthetic references can result in comparable correlation.

\section{Related Work}
The metrics shared task at WMT \cite{ma-etal-2018-results, mathur-etal-2020-results} has played a key role in the development and evaluation of automatic metrics. The annual data collected from WMT’s comprehensive human evaluation of the translation task provides an ideal foundation for assessing these metrics. In this event, metrics are ranked based on the correlation calculated by comparing their scores to human ratings.

Machine translation systems that are compared typically come from the same evaluation campaign. An exception to this is the study by \citet{graham-etal-2014-machine}, who study longitudinal improvements in machine translation quality from 2007--2012 with human assessments, finding that translation quality of the top submissions to the WMT shared task indeed rose significantly.
Our study is similarly longitudinal, but our data stems from a single commercial system, and we do not perform our own human evaluation, but instead assume that improvements over time have been validated by company-internal human A/B testing.

Instead of evaluating metrics through comparison with human judgement, \citet{moghe-etal-2023-extrinsic} explored a complementary approach by correlating metrics with the outcome of downstream tasks. Similarly, our study does not use human judgment directly; instead, we evaluate metrics based on their preference for newer MT outputs.

\section{Methods}

We turn next to describe our data and the metrics we use.
\subsection{Data}
\label{sec:data}

The original corpus contains sentences in English from Abstract Meaning Representation (AMR) Annotation Release 2.0 \cite{knight_abstract_2017}, along with their German, Italian, Spanish, and Chinese translations developed by \citet{damonte-cohen-2018-cross} and released by LDC in \citet{damonte_abstract_2020}. This corpus contains 1371 sentences per language. The source sentences were mainly drawn from content gathered in the news domain.

Translations\footnote{Due to the origin of the translations, the data we use is to be licensed by the Linguistic Data Consortium. If you would like to use this dataset for your research, please contact the authors. The collection of the data continues.} are gathered weekly from May 2018 -- March 2024 using Google Translate from each of the five languages to the other four languages. Early experiments revealed that for English$\rightarrow$Spanish, there was a substantial similarity between professional translations and those generated by the earliest systems (details in Appendix \ref{appendix:es}). Consequently, Spanish was removed from further investigation, reducing the number of language pairs to 12. As minimal variation was observed between consecutive weeks, we subsample for the following analysis, with consecutive systems being approximately one month apart. After removing duplicates (systems receiving identical scores across all metrics), we retained 56--63 systems per language pair.

\begin{figure*}[ht]
    \centering
    \includegraphics[width=0.9\linewidth]{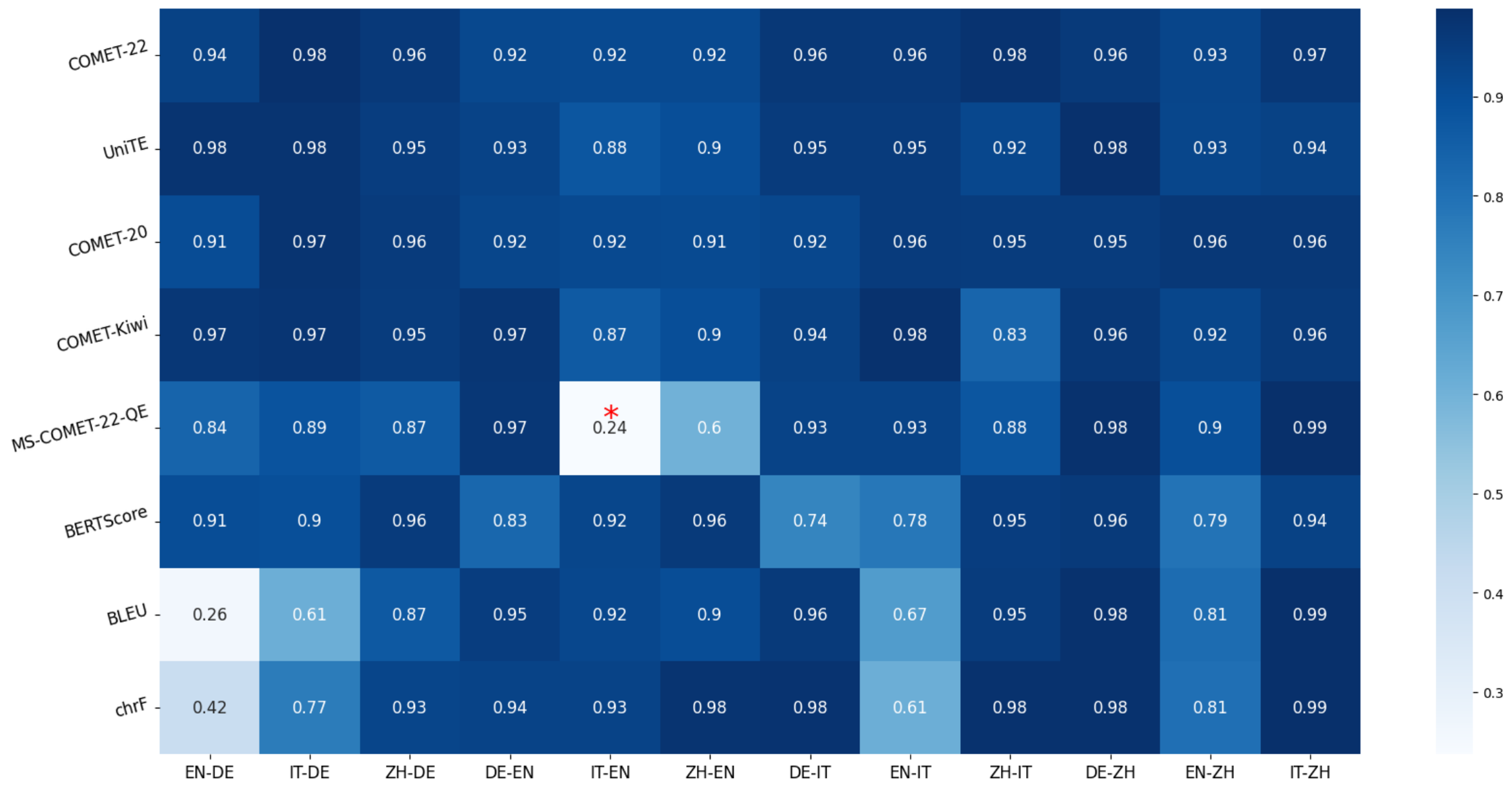}
    \caption{The Spearman correlation measures the relationship between metric score rankings and time rankings for MT systems. A positive correlation indicates an upward trend, with a higher correlation indicating a stronger trend. A red star indicates lack of statistical significance (p-value $>$ 0.05).}
    \label{fig:spearman}
\end{figure*}

\subsection{Metrics}
In this section, we outline the three types of metrics used in this study.
\subsubsection{Surface-level overlap}
\textbf{BLEU} \citep{papineni-etal-2002-bleu} measures n-grams overlap between the translation and its reference. We use \emph{corpus\_bleu} in SacreBLEU \cite{post-2018-call}.

\noindent \textbf{chrF} \citep{popovic-2015-chrf} assesses the overlap between the characters of the translation and the reference. We use \emph{corpus\_chrf} in SacreBLEU.

\subsubsection{Embedding based}
\textbf{BERTScore} \citep{Zhang*2020BERTScore:} derives contextual embeddings from BERT \citep{devlin-etal-2019-bert} models and computes cosine similarity between embeddings of the translation and the reference. We use the F1 score without TF-IDF weighting.

\subsubsection{Trained with human judgements}
\textbf{COMET-20} \cite{rei-etal-2020-comet} is trained on top of XLM-R \cite{conneau-etal-2020-unsupervised} using Direct Assessments (DA) from WMT17 to WMT19. We use wmt20-comet-da.

\noindent\textbf{UniTE} \citep{wan-etal-2022-alibaba, wan-etal-2022-unite} is capable of evaluating translation outputs in source-only, reference-only, and source-reference-combined assessment scenarios. We use unite-mup.

\noindent\textbf{COMET-22} \citep{rei-etal-2022-comet} is the current default model in COMET and trained on DA from WMT17 to WMT20. We use wmt22-comet-da.

\noindent\textbf{COMET-Kiwi} \citep{rei-etal-2022-cometkiwi} is a reference-free metric trained using DA from WMT17 to WMT20, and DA from the MLQE-PE corpus. We use wmt22-cometkiwi-da.

\noindent\textbf{MS-COMET-QE-22} \citep{kocmi-etal-2022-ms} is a reference-free metric, extending COMET by Microsoft Research with proprietary data.

\begin{table*}[ht]
\centering
\begin{tabular}{lcccccc}
\hline\hline
 & All & Into EN & From EN & Into DE & Into IT & Into ZH \\
\hline
COMET-22 & \textbf{73.9} & 66.6 & 71.6 & 76.4 & \textbf{79.4} & 72.6 \\
COMET-Kiwi & 73.4 & 72.1 & 73.9 & 74.8 & 75.3 & 71.4 \\
UniTE & 73.2 & 66.5 & 73.7 & \textbf{77.1} & 75.0 & 73.9 \\
COMET-20 & 72.5 & 66.1 & \textbf{74.6} & 74.3 & 74.0 & 74.9 \\
chrF & 71.4 & \textbf{74.5} & 57.8 & 60.4 & 76.5 & 74.6 \\
MS-COMET-22-QE & 69.9 & 57.4 & 68.1 & 68.8 & 73.9 & \textbf{78.6} \\
BLEU & 68.2 & 71.7 & 57.3 & 56.3 & 68.9 & 76.4 \\
BERTScore & 68.0 & 65.4 & 62.2 & 68.8 & 69.0 & 68.6 \\
\hline\hline
\end{tabular}
\caption{Accuracy for ranking system pairs. Column “All” shows the results for all system pairs. Each following column evaluates accuracy over a subset of systems. Rows are sorted by the accuracy over all system pairs.}
\label{tab:subset}
\end{table*}

\section{Results}
\subsection{How do metric scores change over time?}
\label{subsec:over_time}
While it is reasonable to expect that systems improve over time, how metric scores will reflect these improvements remains unclear. To investigate this, we visualize how metric scores vary over time for individual language pairs in Appendix \ref{appendix:over_time}. In general, upward trends are evident for the metrics across the language pairs. 

We use Spearman correlation to measure whether the upward trends are consistent. Metrics with higher correlation are deemed more reliable, as they better reflect the overall ranking of the systems. As illustrated in Figure \ref{fig:spearman}, COMET-22, UniTE, COMET-20, and COMET-Kiwi consistently demonstrate high correlation across the language pairs. Among the remaining four metrics, we notice low correlations in specific language pairs, like BLEU and chrF in English$\rightarrow$German or MS-COMET-22-QE in Italian$\rightarrow$English.

\begin{figure*}[t]
    \centering
    \includegraphics[width=0.9\textwidth]{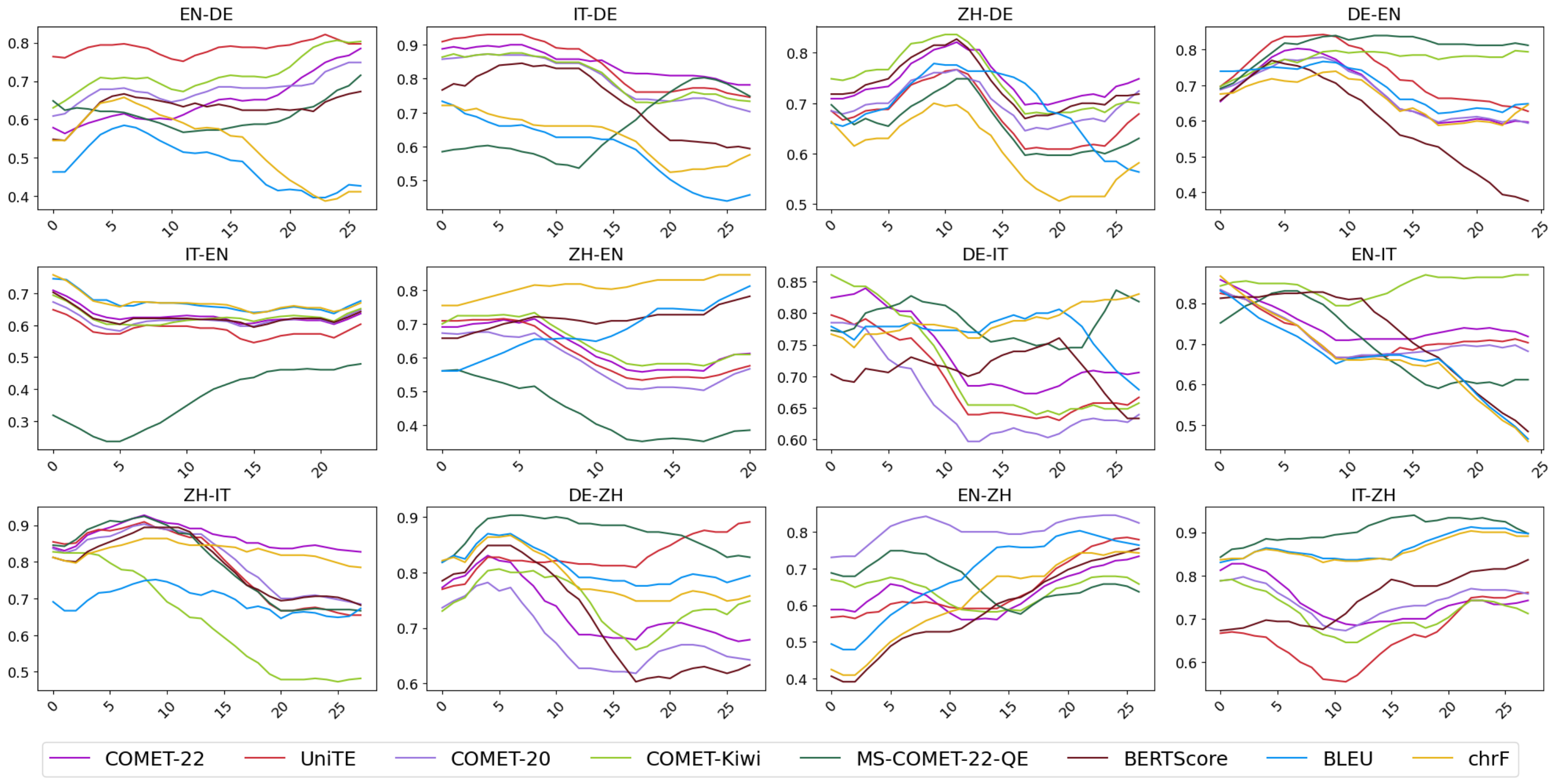}
    \caption{Accuracy over a rolling window of 36 systems. The x-axis represents the index of the starting system, with systems ordered chronologically from the earliest to the most recent. The x-axis scale may vary due to differing numbers of systems, as discussed in Section \ref{sec:data}.}
    \label{fig:roll}
\end{figure*}
\begin{figure*}[t]

    \centering
    \includegraphics[width=0.9\textwidth]{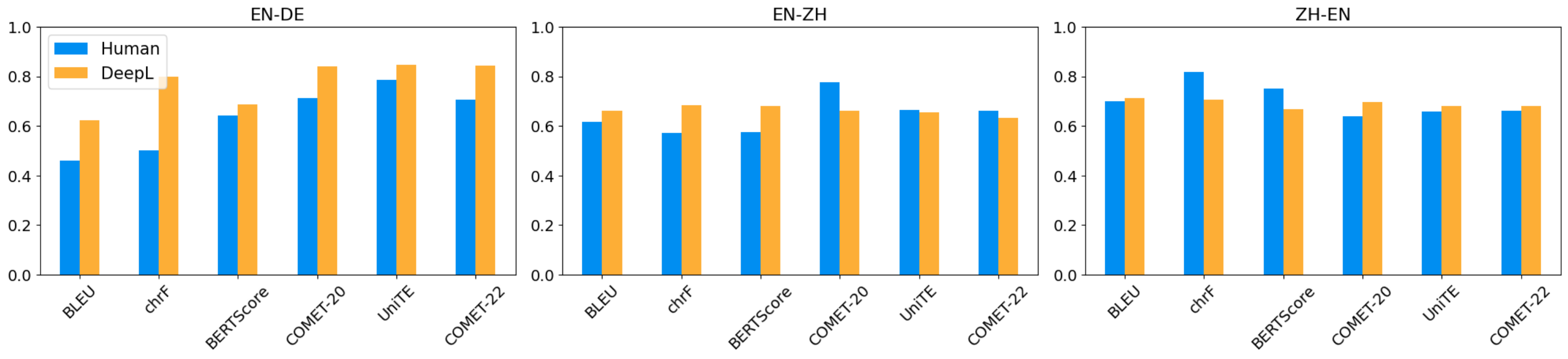}
    \caption[Accuracy across three language pairs using either human or synthetic references.]{Accuracy across three language pairs using either human or synthetic references. The two reference-free metrics are not included as they will not be influenced by reference.}
    \label{fig:deepl}
\end{figure*}
\subsection{How good can the metrics rank incremental systems accurately?}
\label{subsec:acc}
In this section, we evaluate metrics in a common scenario \citep{mathur-etal-2020-tangled}: ranking a pair of systems. As we assume newer systems are better than old  ones, accuracy \cite{kocmi-etal-2021-ship} is adopted as follows. For each system pair, we calculate the difference of the metric scores (metric$\Delta$) and the difference in time (time$\Delta$). Accuracy for a specific metric is calculated as the ratio of rank agreements between metric and time deltas to the total number of comparisons:
\begin{equation*}
    \text{Accuracy} =  \frac{|\text{sign(metric$\Delta$) == sign(time$\Delta$) }| }{|\text{all system pairs}|}
\end{equation*}

Since the systems span from 2018 to 2024, those separated by a substantial time interval might exhibit considerable quality gaps, potentially resulting in an overestimate of metric reliability \cite{mathur-etal-2020-tangled}. Consequently, we only pair systems with a gap of less than a year. Even within such a timeframe, substantial improvements in quality are possible \cite{caswell2020google}. 

Table \ref{tab:subset} shows that trained metrics generally outperform non-trained metrics. For all system pairs, COMET-22 achieves the highest accuracy, followed by COMET-Kiwi. In contrast, MS-COMET-QE-22 struggles to attain high accuracy except for into Chinese. Among surface-level metrics, chrF outperforms BLEU, reflecting results in previous studies \cite{kocmi-etal-2021-ship}, and achieves the highest accuracy for into English. We also examine performance for individual language pairs. Trained metrics exhibit high accuracy, yet no single metric excels across all pairs. More details in Appendix~\ref{appendix:acc}.

\subsection{Does the reliability of metrics depend on the quality of the systems evaluated?}

As mentioned in Section \ref{sec:intro}, metric reliability may decline as the quality of evaluated systems improves \cite{ma-etal-2019-results}. However, the limited number of MT systems made it difficult to fully confirm this trend \cite{mathur-etal-2020-tangled}. We revisit this issue using a larger sample of MT systems. Following the approach of \citet{mathur-etal-2020-tangled}, we implement a rolling window of N systems, transitioning from the earliest to the most recent ones. 

Using accuracy as explained in Section \ref{subsec:acc}, we conduct tests with N varying from 24 to 40. Figure \ref{fig:roll} illustrates the results for N = 36, representing the identified scenarios. Different metrics display varying trends. For instance, in English$\rightarrow$German, trained metrics show an upward trend, while surface-level metrics show a downward trend. A downward trend is most common, with each metric showing a clear decline across 7 or more language pairs. However, we also observe upward or relatively flat trends in the remaining language pairs.

\subsection{Can synthetic references serve as an alternative to human references?}
We generate synthetic references for three language pairs using another commercial MT system, DeepL, and examine their impact on metric evaluation. As depicted in Figure \ref{fig:deepl}, we observe that for English$\rightarrow$German, all metrics achieve a higher accuracy, while for the remaining language pairs, there are some drops. Overall, synthetic references lead to a comparable accuracy for the three language pairs we investigate, suggesting that they can be used when human references are unavailable.

\section{Conclusion}
Our dataset, covering 12 language pairs with at least 56 machine translations each, surpasses previous datasets that typically included only 3 pairs with around 15 machine translations each. Based on the assumption that newer translations of a regularly updated commercial system tend to be of a higher quality, we apply the dataset to revisit prior findings on MT metrics. We provide larger-scale evidence for debated questions such as the relationship between MT quality and metric reliability—issues that previous research was unable to conclusively resolve on smaller datasets. Additionally, the systems are incremental (a baseline compared to improvements developed by the same group), reflecting the most common use case of the metrics. We encourage the use of our dataset for future investigations into MT metrics or the development of MT quality over time.

\section*{Limitations}
Our study bases on the assumption that newer systems of Google Translate outperform older ones due to the quality assurance measures, including human testing, taken before deployment. Although this is a reasonable belief, it might not always be true. 

Recently, LLM-based evaluators have demonstrated great performance in evaluating MT systems. However, we have not included any LLM-based evaluators in this study because it would be costly to experiment with our extensive dataset. 

\section*{Acknowledgments}
We thank Mengyu Wang, Pinzhen Chen, Yftah Ziser, Zheng Zhao and the anonymous reviewers for their comments.
Rico Sennrich acknowledges funding by the Swiss National Science Foundation (project MUTAMUR; no.~213976).

\bibliography{anthology,custom}

\clearpage

\onecolumn

\appendix
\appendixpage

\section{Metric scores for English \texorpdfstring{$\rightarrow$}{→} Spanish translations}

\label{appendix:es}
\begin{figure*}[ht]
    \centering
    \subfloat[BLEU]{\includegraphics[width=0.45\textwidth]{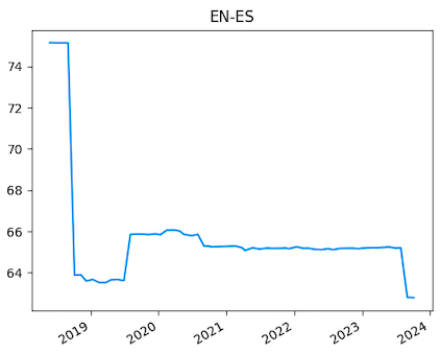}\label{fig:a}}\hfill
    \subfloat[chrF]{\includegraphics[width=0.45\textwidth]{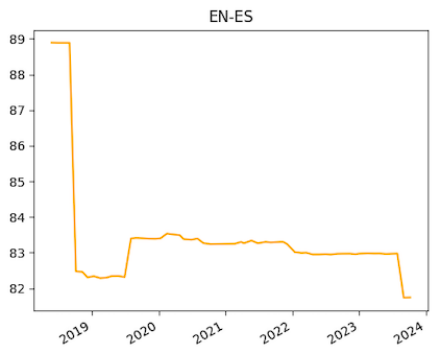}\label{fig:b}}
    
    \medskip
    
    \subfloat[BERTScore]{\includegraphics[width=0.45\textwidth]{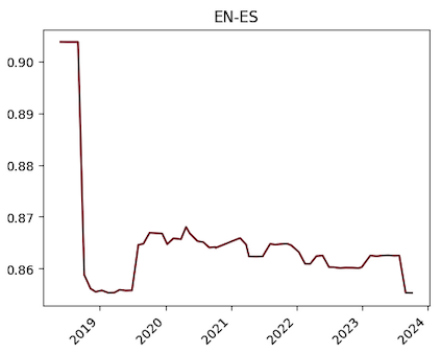}\label{fig:c}}\hfill
    \subfloat[COMET-22]{\includegraphics[width=0.45\textwidth]{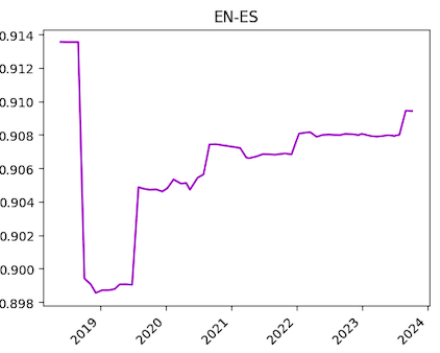}\label{fig:d}}
    \caption{The metric scores for English$\rightarrow$Spanish translations. While the earliest system achieved nearly perfect scores, subsequent systems showed a notable decline.}
    \label{fig:es}
\end{figure*}

Figure \ref{fig:es} displays the scores of four different metrics for English$\rightarrow$Spanish translations in our early experiments. Early systems achieved nearly perfect metric scores, whereas later systems displayed markedly lower scores. Upon closer examination of the human translations, we noticed roughly 25\% of them are identical to that of the early systems. This indicates the use of Google Translate in the professional translation.

\section{Metric scores over time}
\label{appendix:over_time}
Figure \ref{fig:score_over_time} illustrates the findings regarding the change of metric scores over time. Generally, upward trends are evident for the metrics across language pairs. Furthermore, these trends sometimes appear as step-like progressions. Based on a visual inspection of the results, we have some interesting findings as follows:
\begin{enumerate}
    \item Although there have been concerns that MT systems were optimized for BLEU, given its longstanding status as the primary evaluation metric, our findings suggest that the upward trends of BLEU are less consistent compared to other metrics. This observation might provide implicit evidence that BLEU is not solely used during system development.
    \item The trajectories of BLEU and chrF exhibit a high degree of similarity, as do the trajectories of COMET-20, COMET-22, COMET-Kiwi, and UniTE. In contrast, BERTScore and MS-COMET-22-QE follow distinct trajectories of their own. These similarities and discrepancies reflect the inherent properties of these metrics. BLEU and chrF both rely on measuring surface-level overlap, while BERTScore is unique in its reliance on contextual embeddings. As for the trained metrics, although they are all trained in a similar manner, MS-COMET-22-QE was trained using entirely different data.
    \item In certain language pairs, the trajectories of certain metrics may experience a downturn. For instance, noticeable troughs are observed for BLEU and chrF in English$\rightarrow$German, Italian$\rightarrow$German, and English$\rightarrow$Italian; for BERTScore in English$\rightarrow$German, German$\rightarrow$Italian, and English$\rightarrow$Italian; and for MS-COMET-22-QE in Italian$\rightarrow$English, Italian$\rightarrow$German, and Chinese$\rightarrow$English. On the other hand, the trajectories of the remaining metrics may occasionally exhibit bumps but do not show clear troughs.
\end{enumerate}

\begin{figure}[hbt!]
    \centering
    \subfloat{\includegraphics[width=0.9\textwidth]{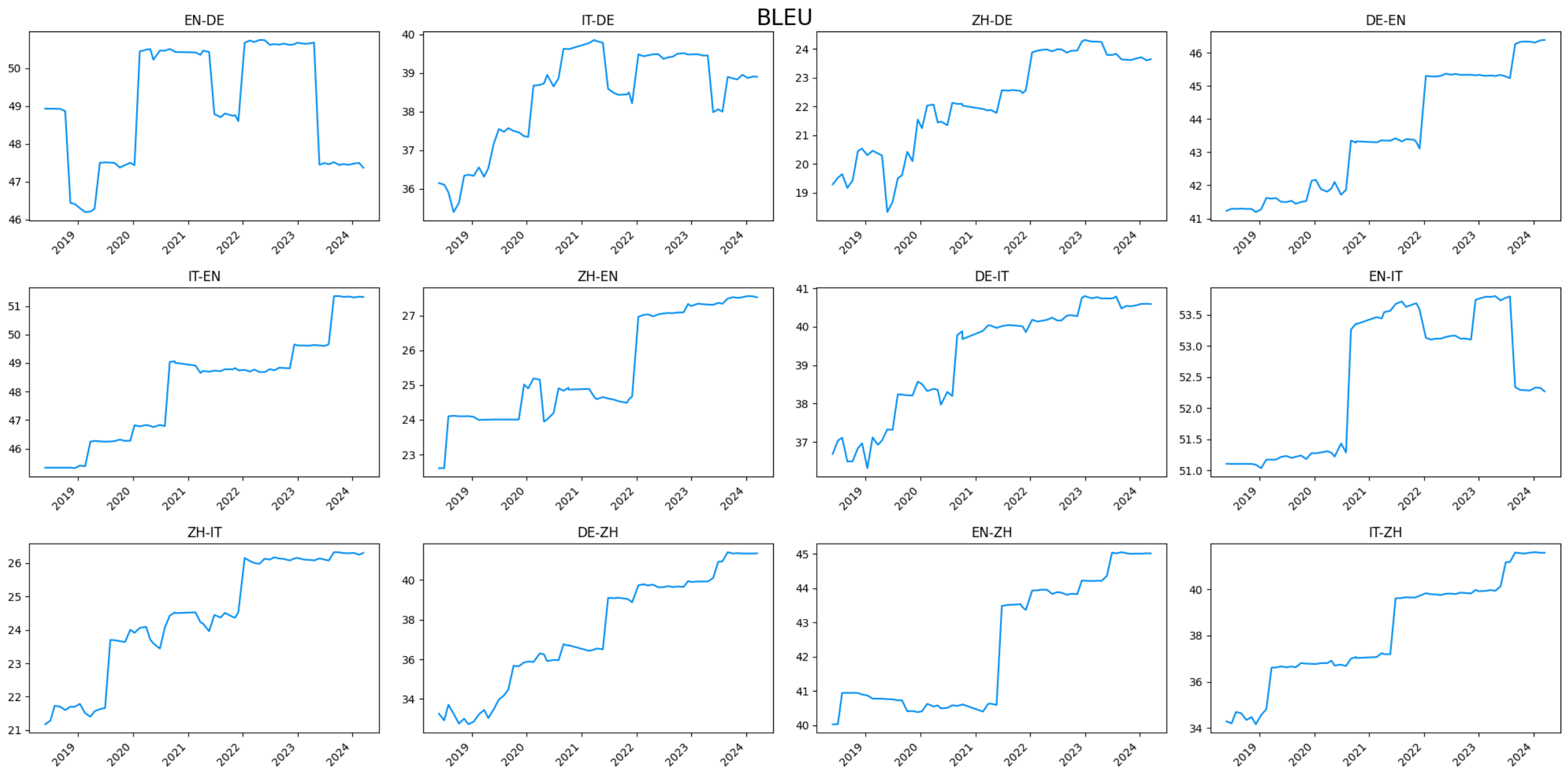}\label{fig:bleu}}
    \vspace{10pt}
    \subfloat{\includegraphics[width=0.9\textwidth]{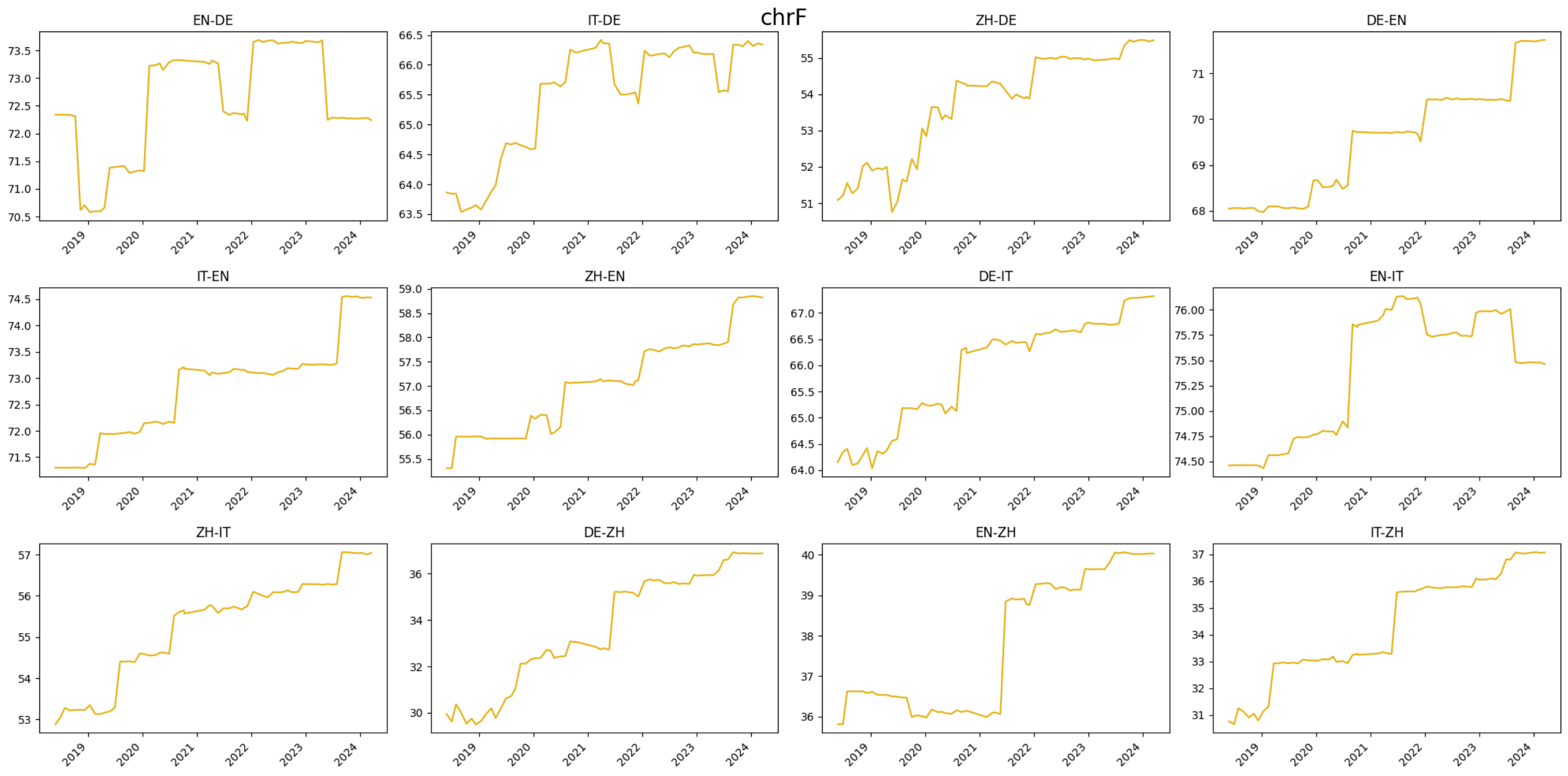}\label{fig:chrf}}
    \vspace{10pt}
    \caption{Metric scores over time.}
    \label{fig:score_over_time}
\end{figure}
\begin{figure}[hbt!]\ContinuedFloat
    \centering
    \subfloat{\includegraphics[width=0.9\textwidth]{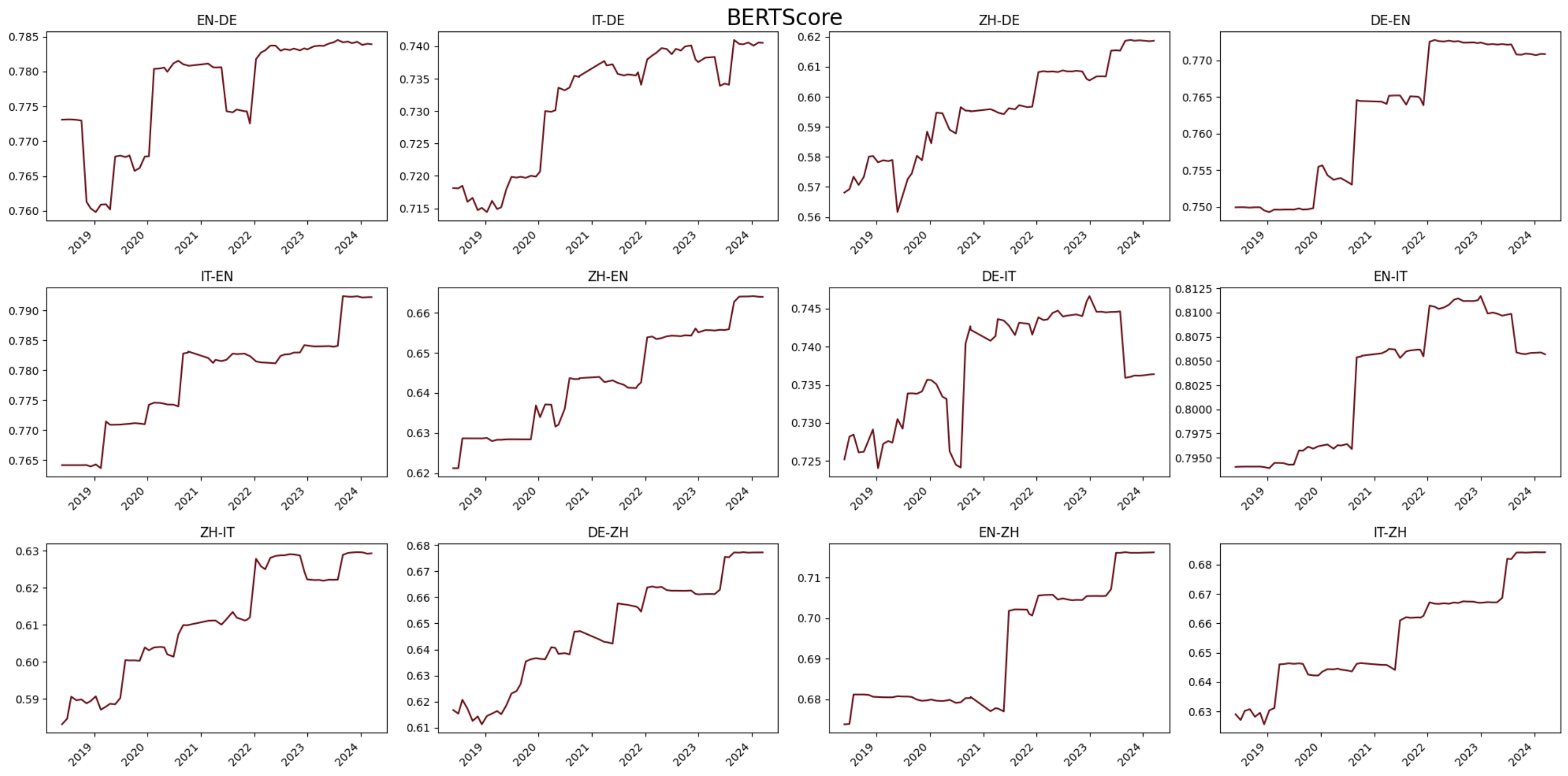}\label{fig:bert}}
    \vspace{10pt}
    \subfloat{\includegraphics[width=0.9\textwidth]{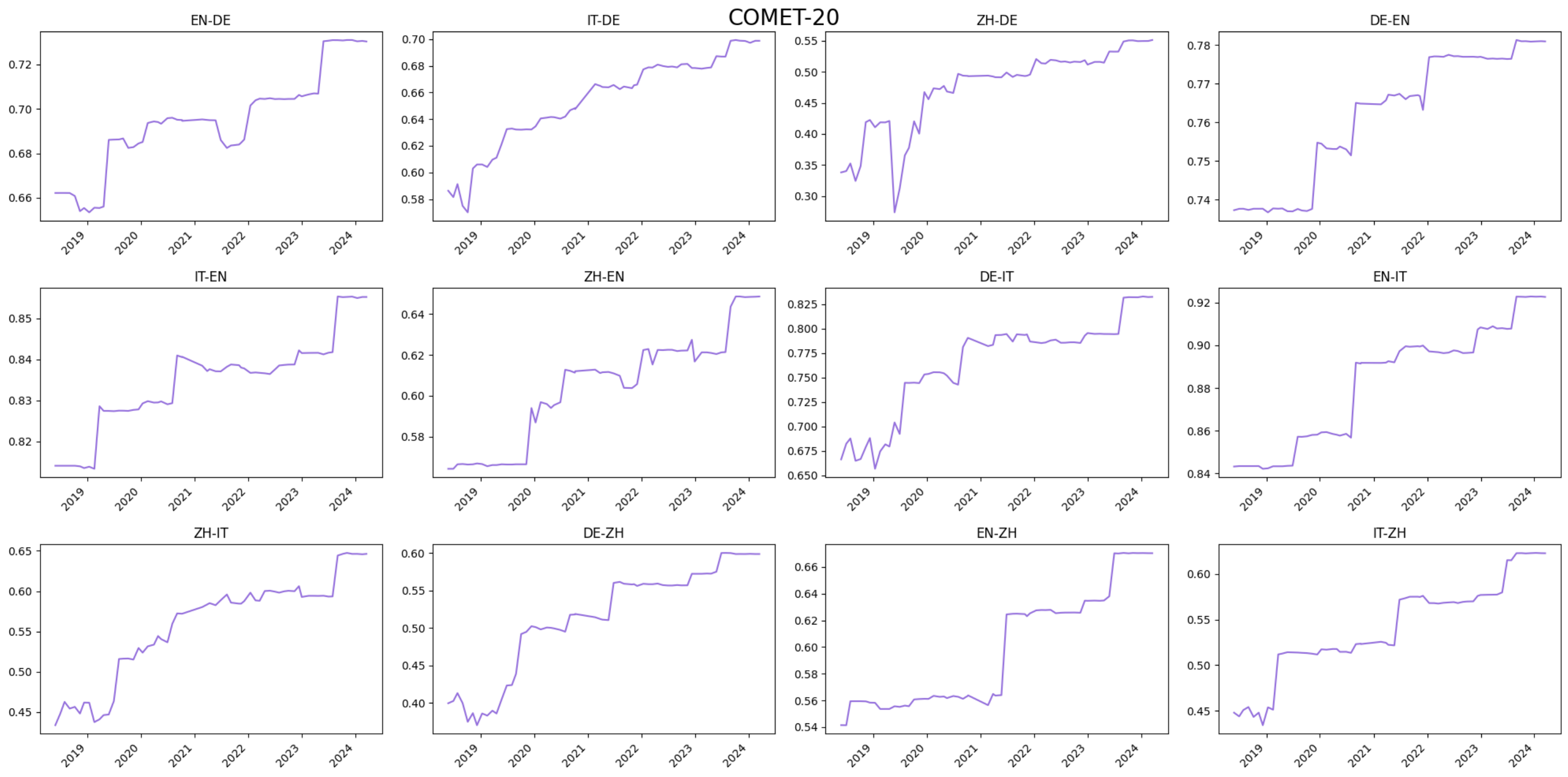}\label{fig:comet20}}
    \vspace{10pt}
    \subfloat{\includegraphics[width=0.9\textwidth]{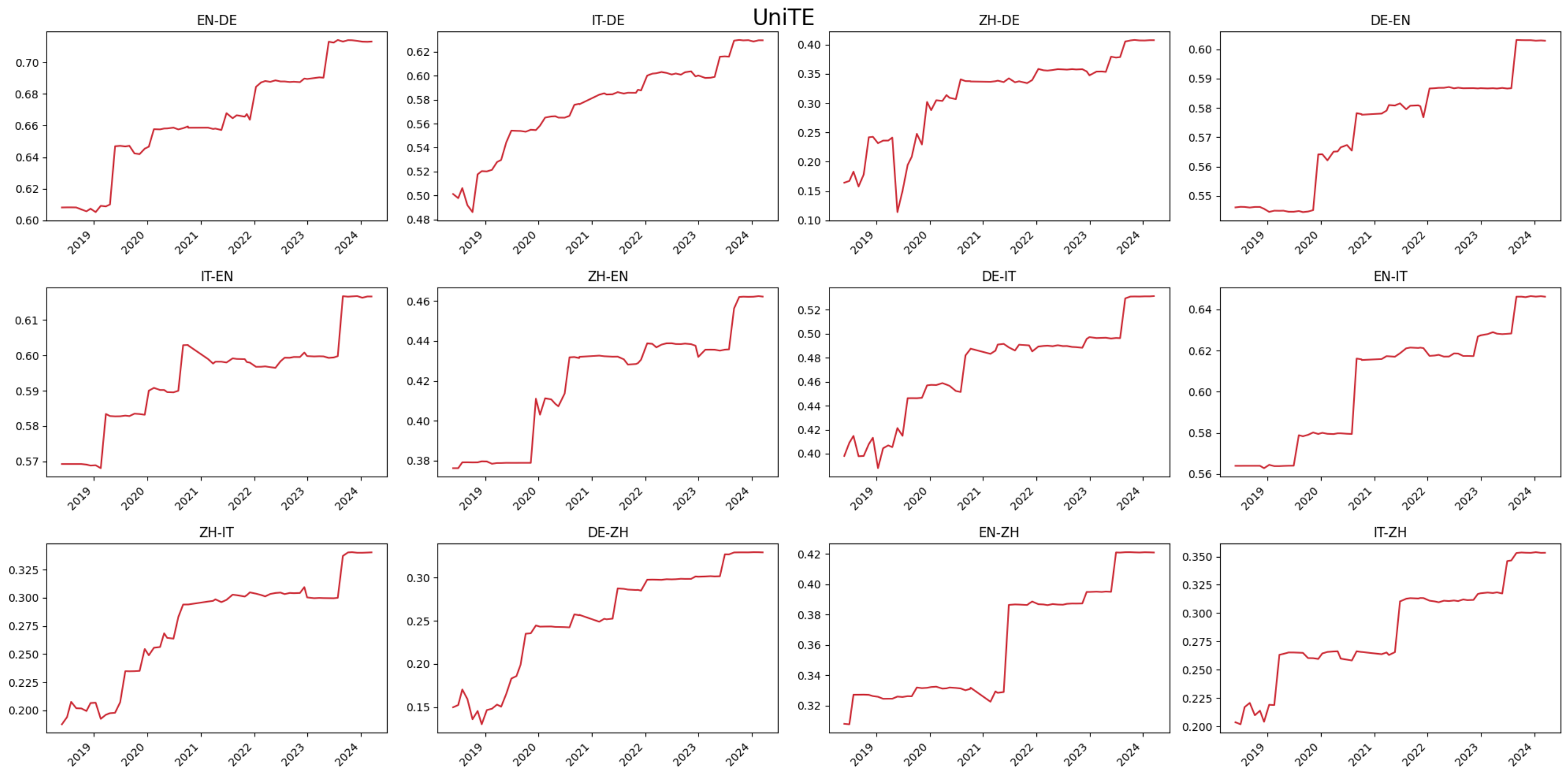}\label{fig:unite}}
    \vspace{10pt}
    \caption{Metric scores over time.}

\end{figure}
\begin{figure}[hbt!]\ContinuedFloat
    \centering
    \subfloat{\includegraphics[width=0.9\textwidth]{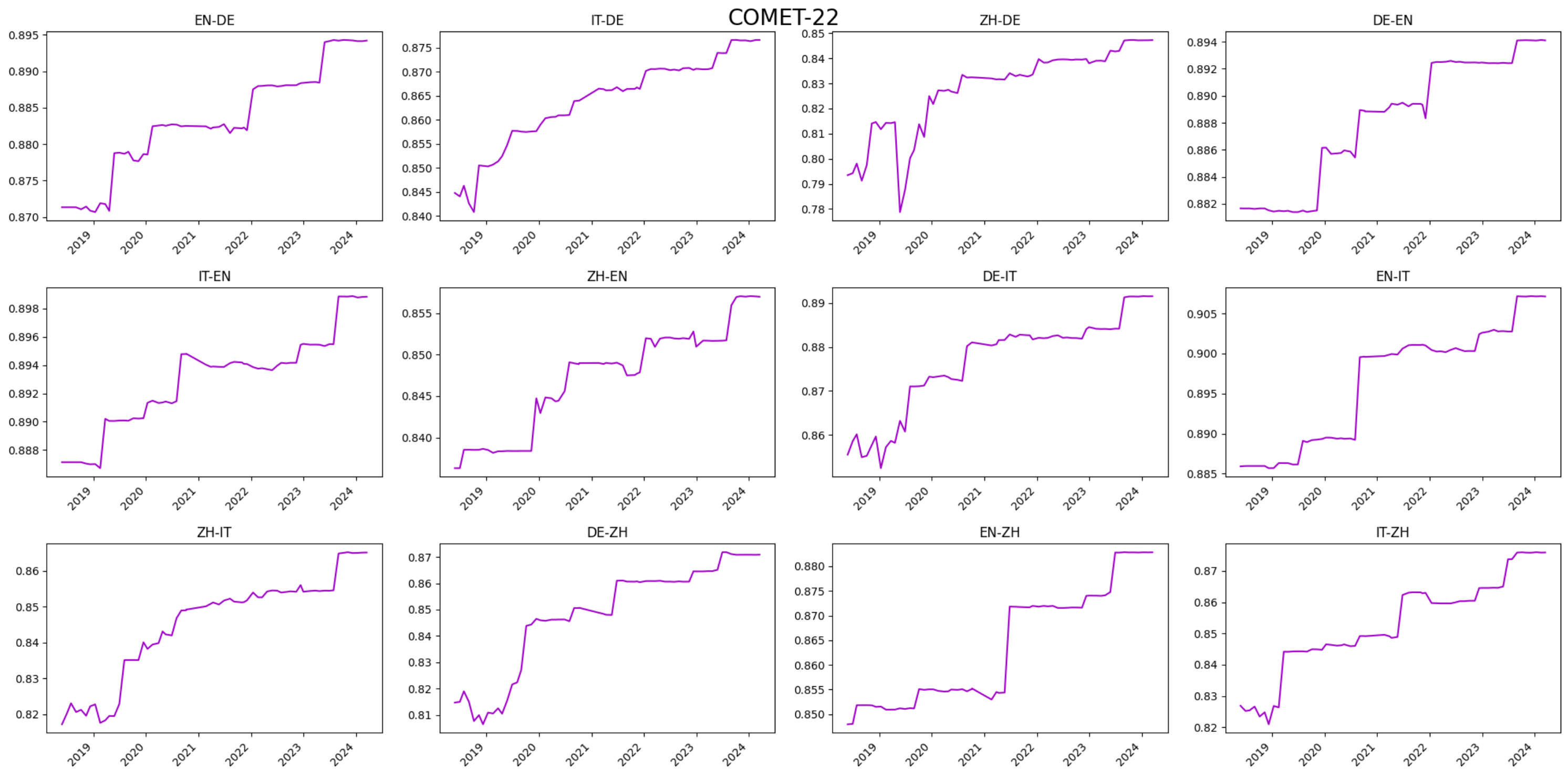}\label{fig:comet22}}
    \vspace{10pt}
    \subfloat{\includegraphics[width=0.9\textwidth]{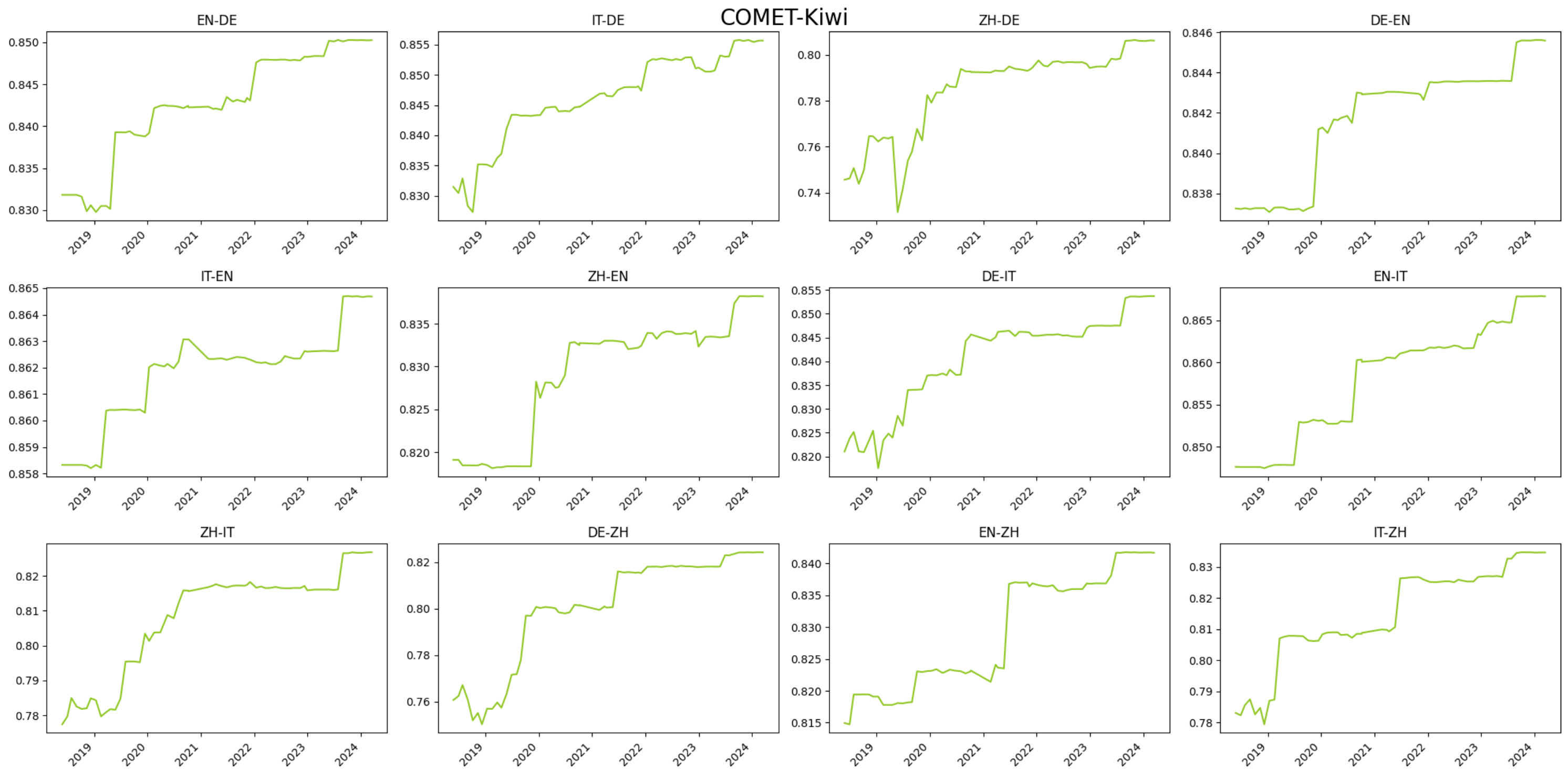}\label{fig:cometkiwi}}
    \vspace{10pt}
    \subfloat{\includegraphics[width=0.9\textwidth]{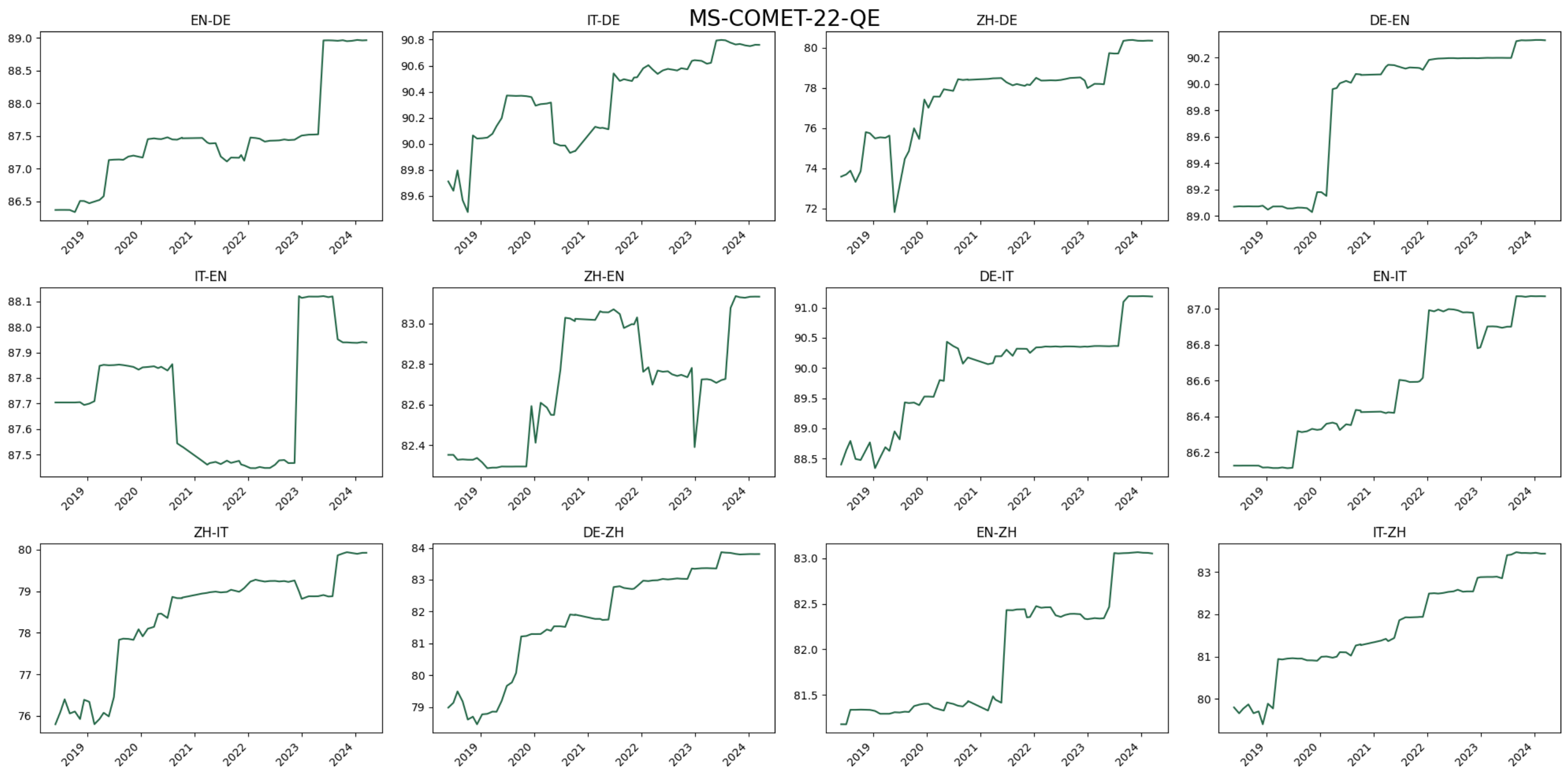}\label{fig:mscomet22qe}}
    \caption{Metric scores over time.}
   
\end{figure}

\newpage
\section{Accuracy across language pairs}
\label{appendix:acc}

\begin{figure}[ht] 
  \centering
  \includegraphics[width=\textwidth]{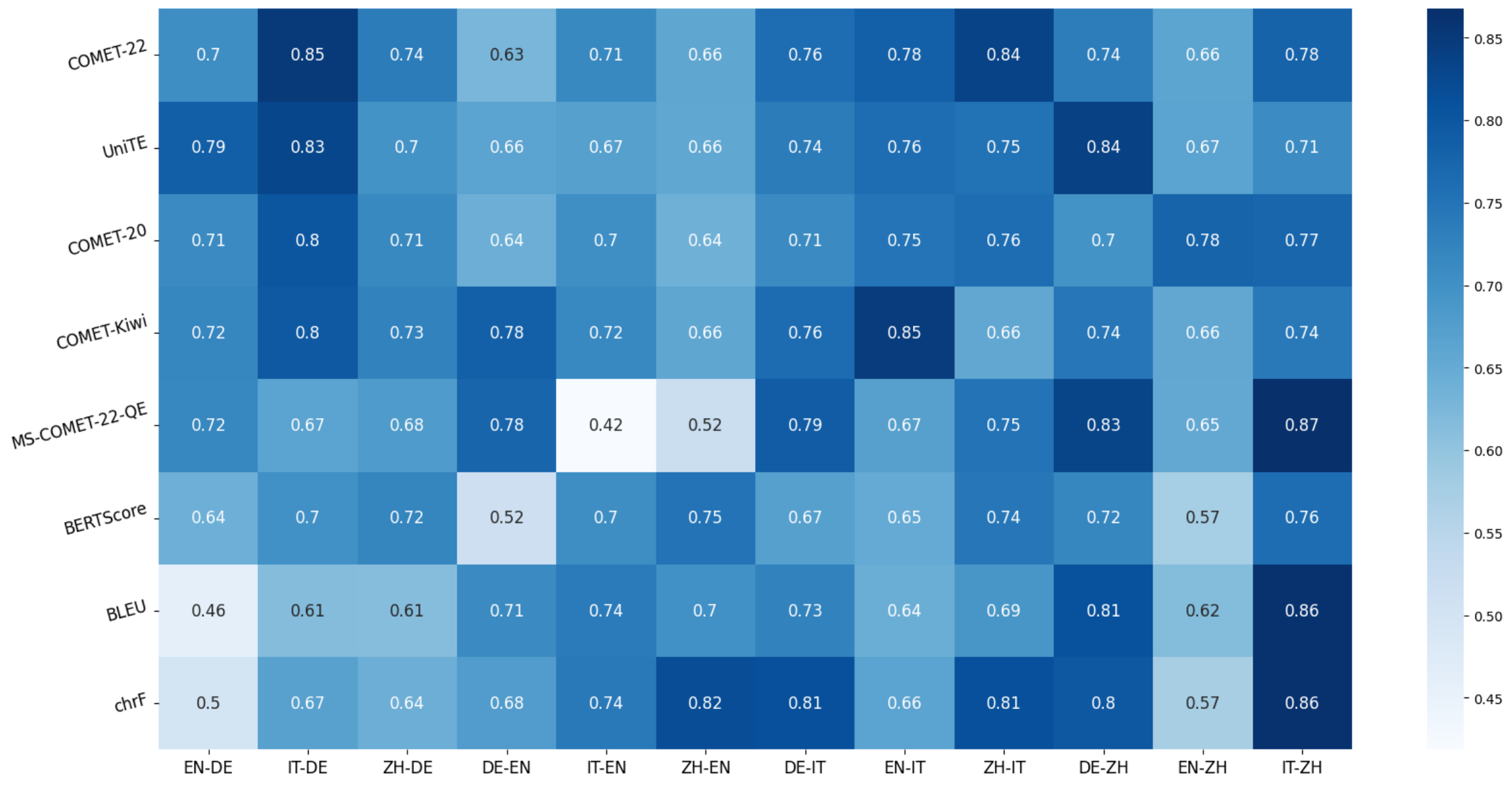} 
  \caption{Accuracy for ranking system pairs across individual language pairs.}
  \label{fig:acc}
\end{figure}

\end{document}